\documentclass[letterpaper, 10 pt, conference]{ieeeconf}  

\IEEEoverridecommandlockouts                              

\usepackage{cite}
\usepackage{amsmath,amssymb,amsfonts}
\usepackage{algorithmic}
\usepackage{graphicx}
\usepackage{textcomp}
\usepackage{xcolor}
\usepackage{hyperref} 
\usepackage{url}

\title{\LARGE \bf
Hybrid Offline–Online Reinforcement Learning for Sensorless, High-Precision Force Regulation in Surgical Robotic Grasping
}

\author{Edoardo Fazzari$^1$, Omar Mohamed$^1$, Khalfan Hableel$^1$,
Hamdan Alhadhrami$^1$, Cesare Stefanini$^1$
\thanks{Partially funded by MBZUAI under project n. SUF-848115}
\thanks{$^{1}$Authors are with the Dept. of Robotics, Mohamed bin Zayed University of AI, Abu Dhabi, UAE.
        {\tt\small \{edoardo.fazzari, omar.mohamed, khalfan.hableel, hamdan.alhadhrami, cesare.stefanini\}@mbzuai.ac.ae}}
}

\begin{document}

\maketitle
\thispagestyle{empty}
\pagestyle{empty}

\begin{abstract}
Precise grasp force regulation in tendon-driven surgical instruments is fundamentally limited by nonlinear coupling between motor dynamics, transmission compliance, friction, and distal mechanics. Existing solutions typically rely on distal force sensing or analytical compensation, increasing hardware complexity or degrading performance under dynamic motion. We present a sensorless control framework that combines physics-consistent modeling and hybrid reinforcement learning to achieve high-precision distal force regulation in a proximally actuated surgical end-effector.
We develop a first-principles digital twin of the da Vinci Xi grasping mechanism that captures coupled electrical, transmission, and jaw dynamics within a unified differential–algebraic formulation. To safely learn control policies in this stiff and highly nonlinear system, we introduce a three-stage pipeline: (i) a receding-horizon CMA-ES oracle that generates dynamically feasible expert trajectories, (ii) fully offline policy learning via Implicit Q-Learning to ensure stable initialization without unsafe exploration, and (iii) online refinement using Twin Delayed Deep Deterministic Policy Gradient for adaptation to on-policy dynamics. The resulting policy directly maps proximal measurements to motor voltages and requires no distal sensing.
In simulation, the controller maintains grasp force within 1\% of the desired reference during multi-harmonic jaw motion. Hardware experiments demonstrate average force errors below 4\% across diverse trajectories, validating sim-to-real transfer. The learned policy contains approximately 71k parameters and executes at kilohertz rates, enabling real-time deployment.
These results demonstrate that high-fidelity modeling combined with structured offline–online reinforcement learning can recover precise distal force behavior without additional sensing, offering a scalable and mechanically compatible solution for surgical robotic manipulation.

\end{abstract}

\section{INTRODUCTION}

Robotic-assisted laparoscopic surgery has introduced substantial advantages over conventional minimally invasive techniques, including enhanced dexterity, tremor suppression, motion scaling, and high-quality three-dimensional visualization~\cite{nadeem2025robot}. These capabilities have enabled surgeons to perform increasingly complex procedures with improved precision and ergonomics. Despite its clinical success, several fundamental control challenges remain unresolved, particularly at the level of distal instrument interaction.

Among these challenges, accurate regulation of grasping force at the instrument tip is of critical importance~\cite{hao2024development}. Effective grasping is essential during suturing, knot tying, needle handling, and delicate tissue manipulation, all of which require task-dependent modulation of applied force. Excessive grasping force may induce localized ischemia, tissue tearing, or irreversible structural damage, whereas insufficient force can lead to unstable grasping, loss of exposure, or dropped instruments. Achieving robust, precise, and adaptive force control is therefore central to safe and efficient robotic surgery~\cite{westebring2009effect}.

Recent research efforts have addressed grasp force estimation and regulation; however, accurate and stable force control in proximally actuated, tendon-driven surgical end-effectors remains largely unresolved~\cite{do2015new}. The inherent nonlinear coupling between motor dynamics, transmission compliance, frictional effects, and distal mechanics significantly complicates controller design. Classical control approaches often rely on simplified models or indirect force estimation strategies, which can limit performance under dynamic surgical motions~\cite{haidegger2009force}.

In parallel, reinforcement learning (RL) has demonstrated remarkable capability in solving complex control and decision-making problems across robotics, game environments, and even physical experimentation~\cite{fazzari2024controlling, tang2025deep, fazzari2025game}. Despite these advances, the application of deep RL in surgical robotics remains comparatively underexplored~\cite{fazzari2026deep}, particularly for high-precision force regulation tasks subject to strict safety constraints.

In this work, we demonstrate how deep reinforcement learning can substantially improve grasping force control in a proximally actuated surgical end-effector. We develop a high-fidelity digital twin of the da Vinci Xi grasping mechanism that captures the coupled electromechanical and tendon-driven dynamics of the instrument. This digital twin is embedded into a reinforcement learning framework to train policies capable of simultaneously regulating grasping force and tracking complex jaw trajectories that emulate the smooth and rhythmic motions characteristic of surgical manipulation tasks.

In contrast to prior approaches that predominantly address either trajectory tracking or force estimation in isolation, this work targets precise grasp force regulation under dynamic motion conditions, where transmission nonlinearities are most pronounced. Rather than relying solely on analytical compensation or hardware augmentation, we integrate oracle-guided trajectory optimization with offline reinforcement learning and subsequent online fine-tuning. This hybrid pipeline enables the synthesis of a compact, high-performance control policy capable of real-time deployment, while explicitly accounting for the coupled electromechanical and transmission dynamics of the system.

The main contributions of this study are as follows:
\begin{itemize}
    \item We develop a physics-consistent digital twin of the da Vinci Xi end-effector that captures coupled motor, tendon, and distal jaw dynamics, and we formulate it as a reinforcement learning environment.
    \item We introduce a hybrid learning pipeline combining oracle-driven data generation, offline Implicit Q-Learning, and online Twin Delayed Deep Deterministic Policy Gradient fine-tuning to safely and efficiently learn high-performance grasp control policies.
    \item The resulting policy maintains grasping force within 1\% of the desired reference during dynamic motion, representing a substantial improvement over prior approaches reporting errors on the order of 20\%~\cite{zhou2024tension}.
    \item We validate the learned controller through experiments on the physical system, demonstrating real-world transfer of the trained policy.
\end{itemize}

\section{Related Work}
Recent research on cable- and tendon-driven surgical instruments has primarily focused on improving transmission modeling accuracy and compensating nonlinear effects such as friction, backlash, and hysteresis. For tendon–sheath actuated systems, adaptive sliding-mode compensation strategies have been proposed to enhance trajectory tracking performance under nonlinear transmission dynamics~\cite{yang2025modeling}. While these approaches demonstrate improved positional accuracy, they typically rely on simplifying assumptions, including constant sheath curvature and pre-tensioned conditions. Moreover, the reported operating force ranges during experimental validation span approximately 20–50 N, indicating substantial force variability that may exceed the requirements of delicate tissue manipulation and fine surgical tasks.

Parallel efforts have concentrated on distal force estimation in cable-driven mechanisms. Model-based backlash compensation algorithms for cable–pulley systems have been introduced to estimate grasping force at the instrument tip from proximal measurements~\cite{zou2022transmission}. In addition, distal sensing strategies such as miniature tension sensor arrays integrated near the end-effector have been proposed to reduce friction-induced estimation errors~\cite{zhou2024tension}, reporting maximum force estimation errors of 1.67 N. While distal sensing improves observability, such approaches introduce additional hardware complexity and integration constraints within confined surgical geometries. Furthermore, analytical compensation methods generally depend on friction modeling assumptions that may not fully capture drivetrain dynamics under time-varying loading conditions.

These findings collectively highlight limited work on the topic and the intrinsic difficulty of achieving robust and accurate distal force regulation in cable-driven surgical instruments, and motivate alternative strategies that move beyond purely hardware-based sensing or purely analytical compensation. Our method relies solely on model-informed learning and compensation of the transmission dynamics. By avoiding end-effector sensing, the approach preserves instrument compactness, sterilizability, and clinical compatibility, while reducing system complexity and cost. This sensorless strategy demonstrates that high-precision distal force regulation can be realized through intelligent modeling and learning-based compensation alone, offering a scalable and translationally attractive alternative to hardware-dependent solutions.

\section{Methods}
\begin{figure}
    \centering
    \includegraphics[width=0.95\linewidth]{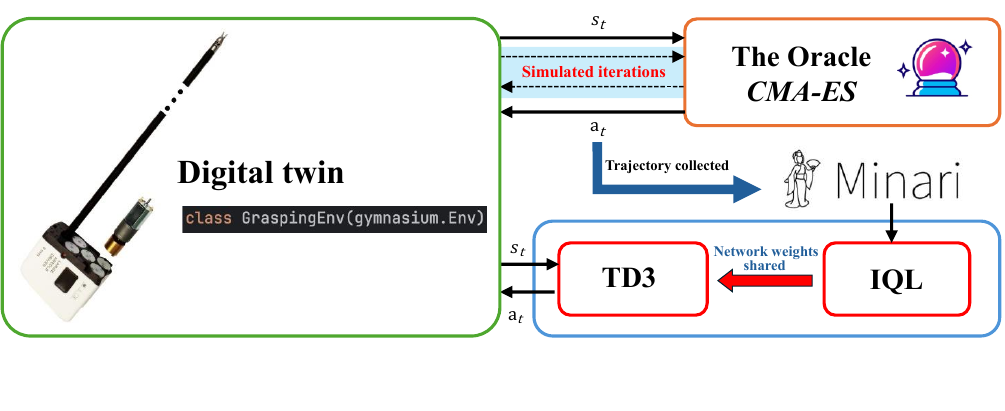}
    \caption{The training pipeline for our final agent based on TD3. }
    \label{fig:pipeline}
\end{figure}

This section describes the digital twin, the learning environment, and the development of the reinforcement learning agent used to solve the task. The overall pipeline of the proposed framework is illustrated in \autoref{fig:pipeline}.

\subsection{The Grasping Digital Twin}
We develop a physics-based digital twin of the da Vinci Xi grasping end-effector that captures the coupled electromechanical dynamics of the dual-motor tendon-driven mechanism. The model is fully derived from first principles and implemented as a coupled differential-algebraic system to preserve dynamic consistency between electrical input, transmission mechanics, tendon compliance, and distal jaw motion.

\paragraph{Modeling Assumptions} The jaws are modeled as rigid bodies with rotational inertia $J_g$ about the distal joint axis. Actuation is realized through stainless-steel wire ropes that are represented as linear axial springs with lumped mass to capture both compliance and dominant inertial effects. Tendon bending stiffness, gearbox backlash, and pulley compliance are neglected, as their contribution is small relative to axial elasticity in the operating regime considered and agonistic-antagonistic configurations. Contact at the jaw tip is assumed rigid, and object compliance is not explicitly modeled; therefore, the computed grasp force corresponds to the internal normal force generated by the symmetric jaw pair. Friction at both the motor and distal joint levels is represented using combined viscous and Coulomb terms. These assumptions retain the dominant dynamics governing grasp force regulation while ensuring computational tractability of the coupled system.

\paragraph{Electrical and Motor Dynamics} Each jaw is actuated by a DC motor with electrical dynamics

\begin{equation}
L \dot{I}_i = V_i - k_v \dot{\theta}_{m_i} - R I_i,
\end{equation}

where $V_i$ is the applied voltage, $I_i$ the armature current, $R$ the internal resistance, $L$ the winding inductance, and $k_v$ the back-electromotive force (EMF) constant. The motor torque is

\begin{equation}
\tau_{m_i} = k_t I_i,
\quad k_t = k_v.
\end{equation}

Rotor and gearbox inertia are lumped into $J_r$. Motor-side friction is modeled as

\begin{equation}
\tau_{f,m_i} = b_m \dot{\theta}_{m_i} + \tau_{c,m} \operatorname{sign}(\dot{\theta}_{m_i}).
\end{equation}

\paragraph{Transmission and Tendon Mechanics}
Motor rotation is transmitted through a gearbox with ratio $\tau$ and efficiency $\eta$ to a spool of radius $r_s$. The tendon elongation for side $i$ is

\begin{equation}
\Delta \ell_i
=
r_s \frac{\theta_{m_i}}{\tau}
-
r_p \theta_g,
\end{equation}

where $r_p$ is the jaw pulley radius and $\theta_g$ the jaw angle.

The tendon stiffness is derived from material properties:

\begin{equation}
k_w = \frac{E A \alpha}{L_w},
\end{equation}

with Young's modulus $E$, effective cross-sectional area $A$ (including fill factor), geometric compliance factor $\alpha$, and tendon length $L_w$.

Unlike quasi-static tendon models, we include lumped tendon mass

\begin{equation}
m_w = \rho A L_w,
\end{equation}

which contributes to the coupled dynamic equilibrium and captures inertia effects at high bandwidth.

The resulting tendon force is

\begin{equation}
F_{s_i} = k_w \Delta \ell_i.
\end{equation}

\paragraph{Jaw and Grasp Force Dynamics}
The jaw rotational dynamics are governed by

\begin{equation}
J_g \ddot{\theta}_g
=
r_p (F_{s_1} - F_{s_2})
-
b_g \dot{\theta}_g
-
\tau_{c,g} \operatorname{sign}(\dot{\theta}_g),
\end{equation}

where $b_g$ and $\tau_{c,g}$ represent distal viscous and Coulomb friction.

The net grasp force at the jaw tip is derived from agonistic-antagonistic force balance mechanics:

\begin{equation}
F_g
=
\frac{r_p}{2r_{\mathrm{tip}}}
(F_{s_1} + F_{s_2}),
\end{equation}

where $r_{\mathrm{tip}}$ is the distance from joint axis to jaw tip. 
This force corresponds to the internal normal force applied by the symmetric jaw pair under rigid contact assumptions.

\paragraph{Coupled Differential-Algebraic Formulation}
Rather than integrating each subsystem independently, the complete mechanism is formulated as a coupled system:

\begin{equation}
\mathbf{A} \mathbf{z} = \mathbf{b}(\mathbf{x}),
\end{equation}

where $\mathbf{z}$ contains accelerations and internal tendon forces,

\[
\mathbf{z} =
[\ddot{\theta}_g,
 \ddot{\theta}_{m_1},
 \ddot{\theta}_{m_2},
 F_{s_1},
 F_{s_2},
 F_g]^\top.
\]

The matrix $\mathbf{A}$ encodes inertia, transmission geometry, tendon stiffness, and force equilibrium constraints. Solving this system at each integration step ensures dynamic consistency between motor inertia, tendon elasticity, and jaw motion.

The full state vector is

\begin{equation}
\mathbf{x} =
[\theta_{m_1}, \dot{\theta}_{m_1},
 \theta_{m_2}, \dot{\theta}_{m_2},
 \theta_g, \dot{\theta}_g,
 I_1, I_2]^\top,
\end{equation}

with inputs $\mathbf{u} = [V_1, V_2]^\top$, yielding

\begin{equation}
\dot{\mathbf{x}} = f(\mathbf{x}, \mathbf{u}).
\end{equation}

\paragraph{Numerical Integration and Stability.}
The system exhibits high stiffness due to the combination of large tendon stiffness $k_w$ and comparatively small inertial terms. To ensure numerical stability under explicit time integration, we employ a fixed timestep of

\begin{equation}
\Delta t = 2 \times 10^{-6} \text{ s}.
\end{equation}

This timestep was selected empirically to guarantee stable simulation without introducing artificial damping. The small step size ensures accurate resolution of high-frequency tendon dynamics and prevents numerical instability arising from stiffness-induced eigenvalues and stick-slip behavior.

\paragraph{Parameter Identification and Physical Consistency}
All physical parameters (material density, Young’s modulus, geometric dimensions, gearbox ratio, motor constants) are derived from manufacturer specifications and measured instrument properties. No artificial tuning terms are introduced. The resulting digital twin preserves energy flow consistency from electrical input to distal mechanical output and provides physically grounded grasp force predictions suitable for control learning.

\subsection{The RL Environment}
The digital twin is embedded into a Gymnasium-compatible reinforcement learning environment in which the agent directly controls the two motor voltages. At each control step, the action consists of the bounded voltage vector $\mathbf{u} = [V_1, V_2]^\top$, with $V_i \in [-V_{\max}, V_{\max}]$ reflecting actuator saturation.

The internal simulator is integrated with a fixed timestep of $\Delta t$~ to ensure numerical stability of the stiff electromechanical system. For each agent action, the simulator advances for $1\,\mathrm{ms}$ of physical time. Consequently, one control decision is applied for exactly $T_c = 1 \times 10^{-3}\,\mathrm{s},$ corresponding to 500 internal integration steps per action. This multi-rate scheme preserves high-fidelity physical dynamics while maintaining a practical control frequency of $1\,\mathrm{kHz}$.

The observation provided to the agent contains motor positions and velocities, motor currents, jaw angle and angular velocity, and the instantaneous grasp force error $F_g - F_{\mathrm{ref}}$. The desired reference force $F_{\mathrm{ref}}$ is set to $12.57~\mathrm{N}$, which corresponds to a stable grasp suitable for surgical needle manipulation. This magnitude provides sufficient normal force to securely hold a needle while avoiding excessive compression that could damage delicate tissue or deform the instrument. Therefore, the episode terminates early when the grasp becomes unstable, i.e., when the grasp force deviates by more than $2.5~\mathrm{N}$ from the reference value; this condition indicates loss of the grasped object (e.g., needle slip or drop) and the task is therefore considered faile

The task requires simultaneous regulation of grasp force and tracking of a time-varying jaw motion. The reward penalizes deviations in both quantities,
\begin{equation}
r_t = -\left(
\lambda_F |F_g - F_{\mathrm{ref}}|
+
\lambda_\theta |\theta_g - \theta_g^{\mathrm{ref}}|
\right).
\end{equation}

The reference trajectories are generated from randomized multi-harmonic functions to emulate the smooth and rhythmic movements characteristic of surgical manipulation tasks~\cite{saveriano2023dynamic}. Two families of trajectories are used, defined as
\begin{equation}
\theta_g^{\mathrm{ref}}(t)
=
\theta_0
+
\sum_{k=1}^{3} A_k \sin\!\big(2\pi f_k t + \phi\big),
\end{equation}
and
\begin{equation}
\theta_g^{\mathrm{ref}}(t)
=
\theta_0
+
\sum_{k=1}^{3} A_k \cos\!\big(2\pi f_k t + \phi\big),
\end{equation}
where $f_1 = f$ is sampled uniformly from $[0.4, 1.0]$~Hz and the higher harmonics are defined as $f_2 = \gamma_2 f$ and $f_3 = \gamma_3 f$ with $\gamma_2 \in [1.5, 2.5)$ and $\gamma_3 \in [2.5, 3.5)$. The amplitudes satisfy $A_2 = \rho_2 A_1$ and $A_3 = \rho_3 A_1$, with $A_1 \in [0.05, 0.15]$~rad and $\rho_2 \in [0.15, 0.35]$, $\rho_3 \in [0.03, 0.12]$; the sign of $A_1, A_2,$ and $A_3$ may be randomized to induce trajectory diversity while preserving smoothness. Finally, the offset is chosen as $\theta_0 = -\sum_{k=1}^{3} A_k$ to ensure bounded motion around the operating configuration.

\subsection{Oracle and Data Collecting}
Preliminary experiments indicated that directly solving the combined force–motion regulation problem using deep reinforcement learning from scratch was prohibitively challenging. The strong nonlinear coupling between tendon elasticity, motor dynamics, and friction, together with stiff system dynamics and strict safety constraints, resulted in unstable learning behavior and frequent task failure during exploration. To overcome this limitation, we developed an oracle controller based on trajectory-level optimization using Covariance Matrix Adaptation Evolution Strategy (CMA-ES)~\cite{hansen1996adapting}.

At each control step, the oracle computes an optimal sequence of motor voltages over a finite horizon of length $H$. The optimization variable is therefore a vector in $\mathbb{R}^{2H}$ containing the voltage pairs applied over the horizon. For a candidate sequence $\mathbf{V} = [V_1^{(0)}, V_2^{(0)}, \dots, V_1^{(H-1)}, V_2^{(H-1)}]^\top$, the digital twin is rolled forward in closed loop for the corresponding physical duration, and the cumulative tracking cost is evaluated.

The objective function minimizes the weighted sum of angle tracking error and force regulation error accumulated over all internal simulation steps within the horizon,
\begin{equation}
\begin{split}
J(\mathbf{V}) &= 
\sum_{k=0}^{H-1} \sum_{j=1}^{N_s} \\
&\quad \left(
w_\theta \left| \theta_g(t_{k,j}) - \theta_g^{\mathrm{ref}}(t_{k,j}) \right|
+
w_F \left| F_g(t_{k,j}) - F_{\mathrm{ref}} \right|
\right).
\end{split}
\end{equation}
where $N_s$ denotes the number of internal integration steps per control action, and $w_\theta$, $w_F$ are weighting coefficients balancing motion tracking and force regulation. The optimization is subject to actuator bounds $V_i \in [-V_{\max}, V_{\max}]$.

CMA-ES is particularly well suited for this setting due to the nonconvexity and nonsmoothness introduced by Coulomb friction and absolute value penalties. The method iteratively samples candidate voltage sequences from a multivariate normal distribution, adapts the covariance matrix based on ranking of solutions, and progressively concentrates the search toward low-cost regions of the control space. To improve efficiency and temporal consistency, the optimizer is warm-started at each timestep by shifting the previously computed optimal sequence forward in time, thereby preserving receding-horizon structure.

Because each function evaluation requires a full rollout of the stiff electromechanical model over the horizon, computational efficiency is critical. For this reason, the cost evaluation routine is implemented using \texttt{Numba}~\cite{lam2015numba} with just-in-time compilation (\texttt{@njit(fastmath=True)}), allowing the inner simulation and cost accumulation loops to execute at near-native performance. This acceleration is essential to accommodate the large number of function evaluations required by CMA-ES (up to several thousand per optimization step).

After convergence, only the first voltage pair of the optimized sequence is applied to the environment, and the procedure is repeated at the next control step, yielding a receding-horizon evolutionary oracle.

Finally, this oracle was used to generate high-quality expert trajectories. Complete episodes were recorded while the oracle interacted with the environment, and the resulting state–action–reward transitions were stored using the \texttt{Minari} dataset framework~\cite{minari}. These oracle-generated episodes form a physically consistent dataset suitable for imitation learning, offline reinforcement learning, and benchmarking purposes.

\subsection{IQL and TD3 Policy Fine-Tuning}
To leverage the oracle-generated dataset consisting in one millions episodes of 3000 steps without further environment interaction, we employ Implicit Q-Learning (IQL)~\cite{kostrikov2021offline}, an offline reinforcement learning algorithm designed to avoid out-of-distribution action evaluation. IQL replaces explicit policy improvement via maximization with an expectile regression value function, thereby stabilizing learning when training exclusively on fixed datasets.

The recorded Minari episodes are aggregated into a unified MDP dataset and used for fully offline training within the \texttt{d3rlpy} framework~\cite{seno2022d3rlpy}. Observation and action spaces are normalized using affine scalers to stabilize optimization and preserve bounded actuator constraints. IQL is trained for $10^6$ gradient updates with batch size $256$, employing actor and critic learning rates of $3 \times 10^{-4}$, temperature parameter $3.0$, and expectile parameter $0.7$. Optimization is performed on GPU to accommodate the scale of the dataset and the number of updates. This stage produces a policy that closely imitates oracle behavior while benefiting from the stability and regularization properties of offline reinforcement learning. 

While IQL provides a strong initialization, offline training alone cannot fully compensate for potential distributional shift between the oracle dataset and the true on-policy visitation distribution. To further improve performance and allow policy adaptation under real interaction dynamics, we perform an additional online fine-tuning stage using Twin Delayed Deep Deterministic Policy Gradient (TD3)~\cite{fujimoto2018addressing}.

The TD3 policy is initialized using the trained IQL model. Specifically, compatible actor network weights are transferred directly, and the critic parameters are reused when dimensionally consistent. Observation and action scalers learned during offline training are retained to preserve input normalization consistency. This initialization strategy ensures that TD3 begins training from a near-expert policy rather than random parameters~\cite{beeson2022improving}.

Online training is conducted for $2 \times 10^5$ environment steps with batch size $1024$, learning rates of $3 \times 10^{-4}$ for both actor and critic, and standard TD3 target smoothing with Gaussian noise ($\sigma = 0.2$, clipped to $0.5$). Additional exploration noise is injected into the deterministic actor to encourage local policy improvement while maintaining grasp stability.

\section{Results}

\subsection{Oracle}
The CMA-ES oracle was first evaluated using empirically selected weights, assigning $w_\theta = 1$ and $w_F = 0.01$ with a planning horizon of $H = 10$. Under this configuration, the controller achieved near-perfect trajectory tracking while maintaining an almost constant grasp force around the desired reference value. The angular motion followed the prescribed trajectory with negligible visible deviation, and the applied force remained tightly regulated throughout the episode. These results confirmed that the evolutionary optimizer is capable of solving the coupled force–motion regulation problem when provided with sufficient planning horizon and appropriate objective weighting.

To systematically characterize the trade-off between force regulation and trajectory tracking, we performed an episode-level Pareto analysis by sweeping the force weight over a logarithmic range while keeping the trajectory weight fixed. For each weight configuration, a full episode was executed and cumulative tracking errors were recorded. The resulting trade-off curve exhibited the characteristic knee structure typical of multi-objective control problems, revealing a well-defined region where further reduction in trajectory error leads to disproportionate degradation in force stability. The selected operating point lies at this knee and represents a balanced compromise between motion fidelity and grasp robustness. After normalization by characteristic magnitudes observed at this operating point, the effective reward weights correspond to $w_\theta = 1$ and $w_F = 0.005$. This normalization improves numerical conditioning while preserving the relative slope of the trade-off curve. Importantly, since policies are learned over a distribution of smooth reference trajectories rather than a single instance, the selected weights reflect the average trade-off geometry of the task family. Although the exact knee location shifts slightly across trajectories due to variations in amplitude and phase, the overall slope remains consistent.

\begin{figure}
    \centering
    \includegraphics[width=0.65\linewidth]{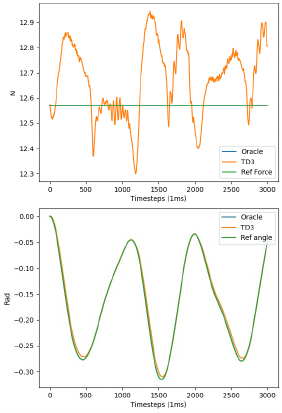}
    \caption{Example of following the same trajectory using the oracle and the fine-tuned TD3.}
    \label{fig:results}
\end{figure}

We additionally investigated the influence of the planning horizon. Increasing the horizon beyond five control steps did not yield measurable improvements in either force stability or tracking accuracy, whereas shorter horizons resulted in visible degradation of performance. Consequently, $H = 5$ was selected as a computationally efficient configuration that preserves near-optimal behavior while substantially reducing the number of function evaluations required by CMA-ES. Under the final parameter selection, the realized trajectory is visually indistinguishable from the reference in \autoref{fig:results}, and the grasp force remains effectively constant with minimal deviation. The mean angular tracking error is approximately 1.17\%, while the average force deviation is in the order of $\mathcal{O}(10^{-4})$, confirming stable regulation under dynamic motion.

Using this finalized oracle configuration, we collected approximately three million transitions across one thousand randomized multi-harmonic trajectories.

\subsection{Fine-tuned TD3}
An example of the behavior of the fine-tuned TD3 policy is shown in \autoref{fig:results}. While the oracle remains marginally more precise, the learned policy achieves strong performance, exhibiting an average force deviation of 1.16\% and a maximum absolute force error of 0.37 N. These results indicate robust regulation, with the controller effectively maintaining a near-constant grasping force under dynamic conditions.  In addition, trajectory tracking performance remains satisfactory, with an average angular tracking error of approximately 5.16\%. Notably, this performance is achieved without internal access to the digital twin during deployment; the policy operates purely as a feedforward mapping from the observed state to motor voltage commands. This compact inference structure enables real-time execution while preserving high-quality force and motion control.

\subsection{Physical Evaluation}

\begin{figure}
    \centering
    \includegraphics[width=0.5\linewidth]{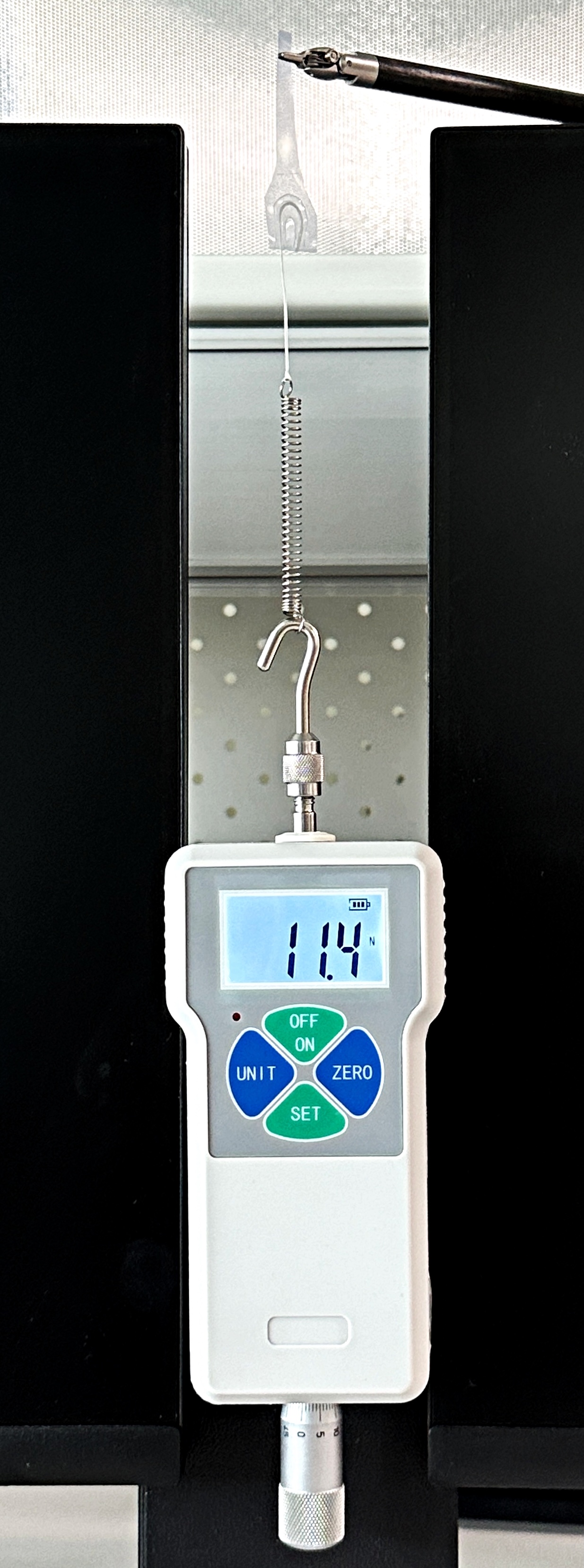}
    \caption{Physical evaluation setup.}
    \label{fig:experiment}
\end{figure}

Experimental evaluation was conducted using the setup shown in \autoref{fig:experiment}. Variations in the grasping force during trajectory execution were measured using a load cell, by monitoring the resistance to slippage of a grasped Mylar leaf. The force required to induce slippage was experimentally correlated with the applied grasping force, demonstrating high repeatability (0.91 ratio).
An elastic element (spring) was used to transmit force from the load cell to the leaf. This configuration enables accurate generation of the desired preload through controlled linear positioning, while also allowing the force to decrease in the event of slippage, thereby reaching a lower equilibrium position. The preload was set using a translation stage driven by a precision graduated screw.
Experiments were performed by progressively increasing the applied force until slippage occurred, and then recording the maximum and minimum force values using a dedicated load cell function. Force inaccuracy was quantified as the ratio between the force error and the average measured force. A total of 10 trajectories were tested, with 5 repetitions each.
The average force error across the entire experimental set was 3.84\%, with the worst-case error equal to 5.67\% and the best result equal to 1.84\%.

A second experimental session was conducted after minor manual optimization of key model parameters (friction coefficients and wire axial stiffness), followed by a new fine-tuning of the TD3 policy. This led to a performance improvement, with minimum and maximum error values of 1.81\% and 5.08\%, respectively, while average was 3.24\%.

\section{Discussion}
The proposed framework demonstrates strong performance in both grasp force regulation and dynamic trajectory tracking, indicating that precise distal manipulation can be achieved without additional sensing hardware or explicit transmission compensation algorithms. By retaining the existing end-effector architecture and regulating only proximal actuation through learned control, the method preserves mechanical simplicity while achieving high-precision force control under motion. This suggests that accurate distal behavior can be recovered through model-informed learning of the coupled electromechanical and transmission dynamics, rather than through hardware augmentation or increasingly complex analytical inversion.

Beyond control accuracy, the compactness of the learned policy makes it particularly suitable for embedded and real-time surgical deployment. The final network contains approximately 71k parameters, corresponding to an estimated memory footprint of 279~KB in FP32 precision, which comfortably satisfies the constraints of modern microcontrollers and surgical control units. Measured inference latency is 0.0386~ms per step on a CPU (MacBook Pro M4), corresponding to an effective inference rate of approximately 26~kHz. Even under conservative assumptions for lower-power embedded processors, the expected latency remains well below 1~ms, enabling high-frequency control with deterministic timing behavior. These results indicate that the proposed approach is not only accurate but also computationally efficient and practically deployable in real-time surgical systems.

\section{Conclusion}
This work presented a sensorless framework for high-precision grasp force regulation in a proximally actuated, tendon-driven surgical end-effector. By developing a physics-consistent digital twin of the coupled electromechanical and transmission dynamics, we formulated grasp control as a reinforcement learning problem while preserving physical interpretability and energy consistency. A structured hybrid learning pipeline combining receding-horizon CMA-ES expert generation, offline Implicit Q-Learning, and online TD3 fine-tuning enabled safe, stable, and data-efficient policy learning in a stiff and highly nonlinear system.

The resulting compact policy achieves near-constant grasp force during dynamic jaw motion in simulation and maintains low force error on the physical platform without distal sensing or explicit transmission compensation. These results demonstrate that accurate distal force behavior can be recovered through model-informed learning of proximal actuation dynamics, avoiding additional hardware complexity while preserving real-time deployability.

Beyond the specific surgical application, this study highlights a general paradigm for controlling coupled, compliant robotic mechanisms: combining high-fidelity modeling with structured offline–online reinforcement learning can bridge the gap between analytical control and purely data-driven approaches. Future work will investigate robustness to varying object compliance, long-term transmission changes, and extension to multi-contact manipulation tasks in realistic surgical scenarios.

\end{document}